# Feedback and Control of Dynamics and Robotics using Augmented Reality

Elijah Wyckoff[1], Ronan Reza[2], Fernando Moreu[2*]

[1] *Department of Mechanical Engineering, University of New Mexico, Albuquerque, NM 87106, United States*
[2] *Department of Civil, Construction and Environmental Engineering, University of New Mexico, Albuquerque, NM 87106, United States*
[*] *Corresponding author, fmoreu@unm.edu, 210 University of New Mexico, Albuquerque, NM 87131-0001.*

**ABSTRACT**
Human-machine interaction (HMI) and human-robot interaction (HRI) can assist structural monitoring and structural dynamics testing in the laboratory and field. In vibratory experimentation, an external force is generated to test dynamic responses of structures. One mode of generating vibration is to use electrodynamic exciters. Manual control is a common way of setting the input of the exciter by the operator. To measure the structural responses to these generated vibrations sensors are attached to the structure. These sensors can be deployed by repeatable robots with high endurance, which require on-the-fly control. If the interface between operators and the controls was augmented, then operators can visualize the experiments, exciter levels, and define robot input with a better awareness of the area of interest. Robots can provide better aid to humans if intelligent on-the-fly control of the robot is: (1) quantified and presented to the human; (2) conducted in real-time for human feedback informed by data. Information provided by the new interface would be used to change the control input based on their understanding of real-time parameters. This research proposes using Augmented Reality (AR) applications to provide humans with sensor feedback and control of actuators and robots. This method improves cognition by allowing the operator to maintain awareness of structures while adjusting conditions accordingly with the assistance of the new real-time interface. One interface application is developed to plot sensor data in addition to voltage, frequency, and duration controls for vibration generation. Two more applications are developed under similar framework, one to control the position of a mediating robot and one to control the frequency of the robot's movement. This paper presents the proposed model for the new control loop and then compares the new approach with a traditional method by measuring time delay in control input and user efficiency.

**Introduction**
The experimental response of structures is quantified by measuring vibrations with sensors. Smart infrastructure wireless sensors are valuable because they are reliable, low-cost, low-power and have fast deployment characteristics [1]. Often, sensor boards are equipped with accelerometers to measure vibrations for structural health monitoring (SHM) [2]. Accelerometer-equipped sensors are especially useful for measuring and detecting low-frequency vibrations [3],[4]. Wireless sensor networks can be built as a collection of sensors and are used for monitoring and assessing vibration risk in structures like historical buildings [5]. For this research accelerometer-equipped low-cost sensors are used, and vibrations are generated at low-g values.

Electrodynamic exciters are a common tool for generating vibrations for laboratory experimentation. Sensor failure due to mechanical vibrations and shock is tested prior to field deployment with a frame that includes three electromagnetic actuators for mechanical excitation as well as loudspeakers for acoustic excitation [6]. Control systems for such exciters include sinusoidal signals, adaptive algorithms, signal amplitude adjustment are examined by Čala et. al, who propose their own model for generating sinusoidal sweep and broadband random vibrations using an inverse filter [7]. For this paper, the authors develop a new interface for manual control of exciter voltage and frequency.

Human-robot interaction is an important area that is gaining popularity in dynamics as well. For example, Popa et al. use robots to deploy sensor networks [8]. Past work in HRI has seen development that mimics capabilities that are granted and improved through AR. For example, Prusaczyk et al. develop software to sort products using a Kinova robot equipped with a vision system [9]. Palacios also demonstrated preliminary work in virtual joystick teleoperation of 6DOF arm [10]. Though the control interface does not involve mixed reality it demonstrates development that led to AR/VR robotic control. Physical sensors have been incorporated with robots as well for blind obstacle detection where accelerometers are used for blind detection of obstacles and spatial mapping to avoid collision with a robotic arm [11]. This capability can now be implemented with AR head-mounted devices (HMD). AR has been applied for an application for more intuitive, less training-intensive means of controlling robots than traditional joystick control [12]. This includes moving a holographic digital twin end effector to desired location and previewing the action of the robotic arm; however this method can be cumbersome in adjusting the digital twin correctly and has limited interaction with the real environment. Similarly, hand tracking and manual movement of a digital twin for moving a robotic arm has been demonstrated in a project by ABB robotics [13]. However, the inaccuracy associated with this mode of control is not suitable for operations such as assembly or production as precise movement and placement of objects is difficult when operating this way. Chacko et al. developed a novel AR spatial reference system for mobile ground robots that is suitable for novice end users to effortlessly provide task-specific spatial information to the robot [14]. This includes placing spatial markers to allocate tasks for the robot at specific locations, where the markers may not perfectly align with the real world as placement depends solely on the user. In recent years interest and development of AR-related HRI has grown exponentially, however there still exist knowledge gaps and insufficient capabilities that need to be addressed.

AR is useful as both a visualization tool and a tool for control, which means AR technology can be developed to provide information for intelligent decision making. AR combines digital elements with the real-world. An AR headset can be worn to project holograms into the user's vision by an optical see-through display. The user operates the headset through a series of gestures and voice commands that allow them to interact with the AR elements. For this research, the headset used is the Microsoft HoloLens 2$^{nd}$ generation (HL2). Moreu et al. [15] provide an extensive overview of the headset's properties and capabilities. Universal Windows Platform development for the HoloLens is supported in Unity, a game engine that can be used to develop AR applications. Registration and tracking issues have stalled AR implementation in application to structures and buildings; thus an algorithm has been developed to implement building information modeling to mitigate issues with AR [16]. Researchers have developed an application to augment displacement data collected by sensors where these values are first recorded and stored in an SQL database before the data can be shown in AR [17]. AR allows the developer to create relevant interactable holograms useful for information and control, and this research leverages this technology for vibratory experimentation.

This paper implements AR technology for dynamic experimentation and robotics by allowing researchers to operate control inputs while also monitoring sensor data. There are two applications developed which consolidate multiple tasks into each of the individual interfaces. While running an experiment with this technology, the AR user receives information independent of positioning and gaze focus. Traditional methods of sensor data monitoring include devices with screens displaying information, and manual control of electrodynamic exciters is done with separate devices as well. The new interface combines these elements into one user-friendly environment. The interface uses the LEWIS5 (Low-cost Efficient Wireless Intelligent Sensor) an Arduino-based Metro M4 microcontroller equipped with an accelerometer and WiFi shield. The HoloLens Gen 2 AR headset connects to the board over WiFi, and sensor data is sent and plotted as a hologram in the same interface as the controls which send commands to the exciter from the headset. The accelerometer on the LEWIS5 measures vibration levels, and the exciter is connected to a LEWIS5 as well which receives and interprets the commands for the exciter. The robotic arm is controlled by a MATLAB program that pulls frequency and position values from a database, for which the values are communicated by the AR application. The proposed actuator control application validated through simple experimentation testing the time delay of control compared to a traditional method of manual control. The robotic application is tested by measuring inaccuracy in placement and frequency output.

**Background**

This research is motivated by intelligent human-structure interaction enhanced by sensors and robotics. In the case of experimentation humans have a better sense of reality when aware of the real structural response and the data measured by sensors. Sensors do not inform on all events that humans are aware of by observation, and humans cannot quantify the structural response without the data from sensors. Therefore, experimenters must be aware of the real structure and the vibration data to make informed decisions, which include changing external input. Monitoring data, controlling actuators, controlling, and interacting with robots, and monitoring structures divide the experimenter's attention. A review of AR technology found that humans receive between 80-90% of information through vision [18]. According to that same study, the amount of information humans can receive, and process is limited by their mental capacity, so AR can help reduce the cognitive load. AR has been applied to robot teleoperation to reduce gaze distraction by augmenting live video feed from the robot [19]. Instead of constricting the user's focus to one task, this application of AR allows the user to focus on information from the robot as well as control. AR technology can consolidate information to reduce the human's cognitive load.

There are three applications developed to apply AR to improve human cognition: Actuator control, robot position control, and robot frequency control. The first application is tested as a first-step investigation into control with AR, where time delay is a primary concern as a barrier to effective implementation. The actuator used in this program is an electrodynamic exciter. The exciter is a portable permanent magnet actuator with an integrated power amplifier in its base. The excitation signal from a function generator is sent by BNC connector at the actuator's base. To receive input signal from the HL2 to the exciter, an Arduino board connects to a 3.5mm adapter. Effective interaction between robots and humans is emphasized in the two robot applications. Human-robot interaction is demonstrated in Figure 1a, where the robot used in this research is the Kinova Gen3 shown in Figure 1b. Frequencies are collected which can be interpreted and reproduced by humans who move at their own frequency for interaction with robots and structures. The next section describes the methods implemented to develop these applications.

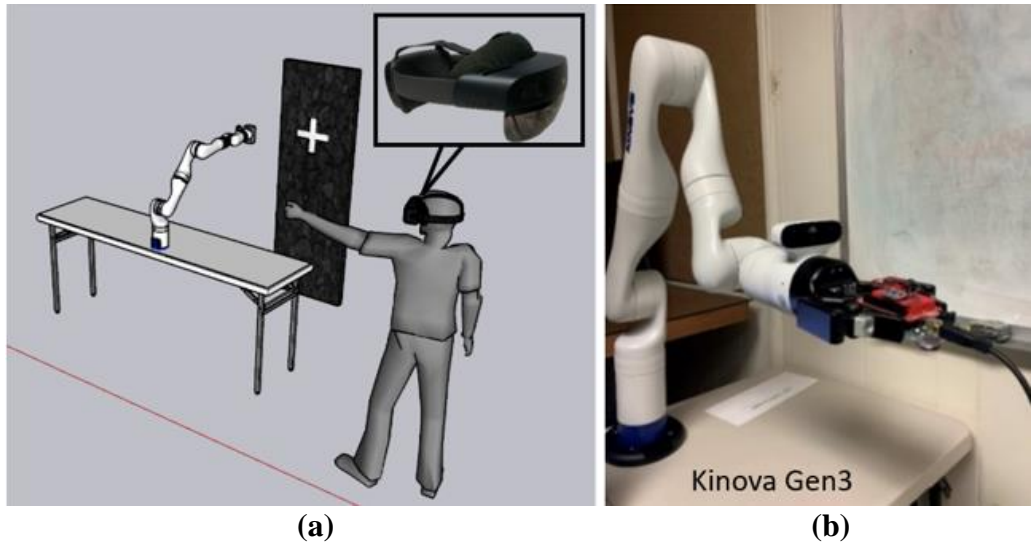

**Fig. 1** Robot arm for experimentation. **(a)** Demonstration of model with HL2; **(b)** the physical arm moving a sensor.

**Methods**
This section describes the hardware and software development applied to achieve high levels of feedback and control. Sensors are used to inform on data that is not obvious to the human eye and showing this data in AR allows the human to monitor both the physical response and the data. AR is leveraged for robotics as it enables on-the-fly control without having to repeat the programming process. However, manual control of the arm even with the aid of AR does not give the human fine control for placement tasks nor does it contain a solution for movement of the base on a dynamic surface. Therefore, this research seeks to provide a solution by augmenting mobile target points set by the user for placement commands while communicating the frequency of any base movement to robot to counteract and stabilize. This section explains the methods implement to achieve these goals.

**Sensing platform**
The sensing platform is developed to collect z-axis (in the direction of G) acceleration wirelessly. This is done with the LEWIS5 sensor, which is built by combining a WiFi shield and microcontroller with a triaxial accelerometer. The sensor connects wirelessly over WiFi but does require a power source that is connected via micro-USB. The fully assembled sensor is labeled in Figure 2. The Metro M4 Express is a 32-bit microcontroller. The Metro M4 microcontroller does not have WiFi built in, so the addition of the WiFi Airlift shield is necessary for wireless connection and data transfer. The triaxial accelerometer used for this project is a highly accurate MMA8451 accelerometer, which measures vibrations in the range of 0.8-1.2 G.

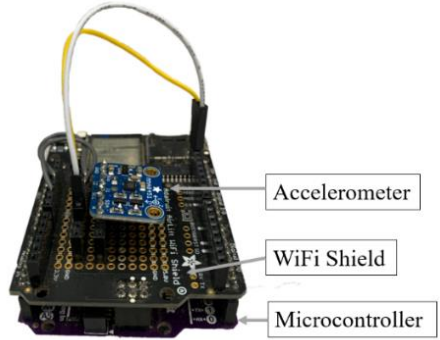

**Fig. 2** LEWIS5 sensor full assembly.

**AR development**
Programming and development of the AR applications is done in Unity version 2019.4.10f1 from Microsoft for AR development. The Mixed Reality Toolkit (MRTK) can be applied to a scene built in Unity to configure the scene for AR. The application is developed for UWP which allows for deployment to the HoloLens 2. The programming platform used is Visual Studio 2019 and any Unity scene programs are written in C#. The sensor programming for the two applications which implement LEWIS5 was performed in the Arduino IDE which allows development for Arduino-compatible boards. The board is setup as a server so it can receive messages from clients connected on the same WiFi network and port.

**Actuator control method**
The actuator control application serves as first-step development into control with AR. Modified code from Timur Kuzhagaliyev [20] is implemented for connecting the HoloLens and Unity to network sockets. The graph of the live data is developed as a scatter plot, which was chosen as the most effective and efficient solution. The graph is developed based on a tutorial from Jasper Flick [21]. Four sliders are created in Unity for the user to define values of voltage, frequency, value multiplier, and duration. The first slider is used to change the value of the exciter's amplitude. Changing the frequency defines the frequency of the sinusoidal signal the exciter generates. The multiplier value makes it possible for the user to increase the other values past the limit induced by the length of the slider. Finally, the fourth slider defines how long the exciter runs. When the sensor receives a message from the HoloLens it calculates the sine function with 256 entries at the voltage setting defined in the HoloLens application by the user.

**Robot control application method**
The target position for the position control application is set based on the placement location that is desired by the human. This defines the robot coordinate system in the same reference frame as the HMD. Thus, it is possible to calculate the position vector necessary for the robot to perform the desired operation. This vector is calculated by the HL2 sensors in meters in the HL2 coordinate system and is sent to a MySQL database where it is permanently stored. As demonstrated in Figure 3, this serves as the connection between the HL2 and the robot as MATLAB pulls the most recent position from the database and runs the code written to operate the position of the robot's end effector. The loop is closed by the user's decision making based on visual feedback of the robot's position.

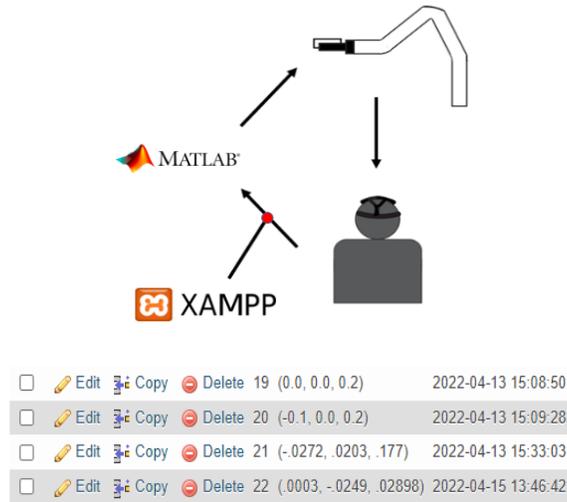

**Fig. 3** Robot control applications methods. Flowchart of robot position control application development.

To control the frequency of the robot movement the Arduino program is written to calculate the frequency of movement along any axis of the accelerometer. Sensor data can be communicated to the robot to offset potentially harmful dynamics, whether through a sensor attached to the base or by human input. Figure 4 shows this interface from the HL2 view where Label 1 indicates the moving base represented by an actuator, and Label 2 is the frequency of the human which is shown in AR and run by the robot.

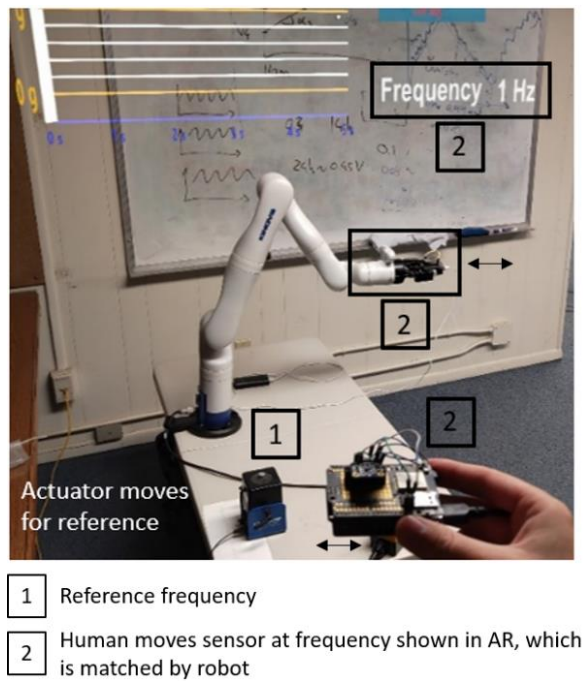

**Fig. 4** Human actuation demonstration, observation and generation of frequency aided by AR.

Arduino FFT library is implemented as it contains a definition for fast Fourier transform in the Arduino IDE. Fast Fourier transform is an optimized algorithm for implementing the discrete Fourier transform (DFT), and the DFT is defined by Equation 1, where N is the length of the filter and k = 0, 1, …, N-1.

$$X_k = \sum_{n=0}^{N-1} x_n * \left[\cos\left(\frac{2\pi k n}{N}\right) - i * \sin\left(\frac{2\pi k n}{N}\right)\right] \tag{1}$$

The code takes a defined number of samples from one direction of accelerometer data and calculates the most dominant frequency in that range using the forward FFT function with Hamming windowing. The Hamming window is given by Equation 2.

$$w(k) = 0.54 - 0.46 * \cos\left(\frac{2\pi k}{N-1}\right) \tag{2}$$

The peak value is calculated every 2.67 seconds and each new value is posted to the database. Simultaneously, the value displays in the HL2 application which is also reading from the database. The user reacts to the exact value they are generating while monitoring the signal of the handheld sensor as well, as demonstrated in the results section.

**Results and Discussion**
This section reports the results of the AR applications developed to enhance human control in dynamics. The three applications are: Actuator control, robot position control, and robot frequency control. The reported results validate the effectiveness of these methods of communication, interaction, and control.

**Actuator control – resulting application**
The result of actuator control development for AR is an application that plots sensor data in the same interface as the electrodynamic actuator controls. This gives the human a full understanding of structural data, which informs decisions to intelligently adjust the dynamic input. The full view of the interface is shown in Figure 5. The application interface consists of four buttons for the sensor graph and four sliders for exciter input with a send button. There are buttons to connect and disconnect from the vibration sensor, and two buttons to start and stop the view of the graph as well. The sliders for exciter control include voltage, frequency, a multiplier, and seconds. The send button sends the current values to the sensor connected to the exciter.

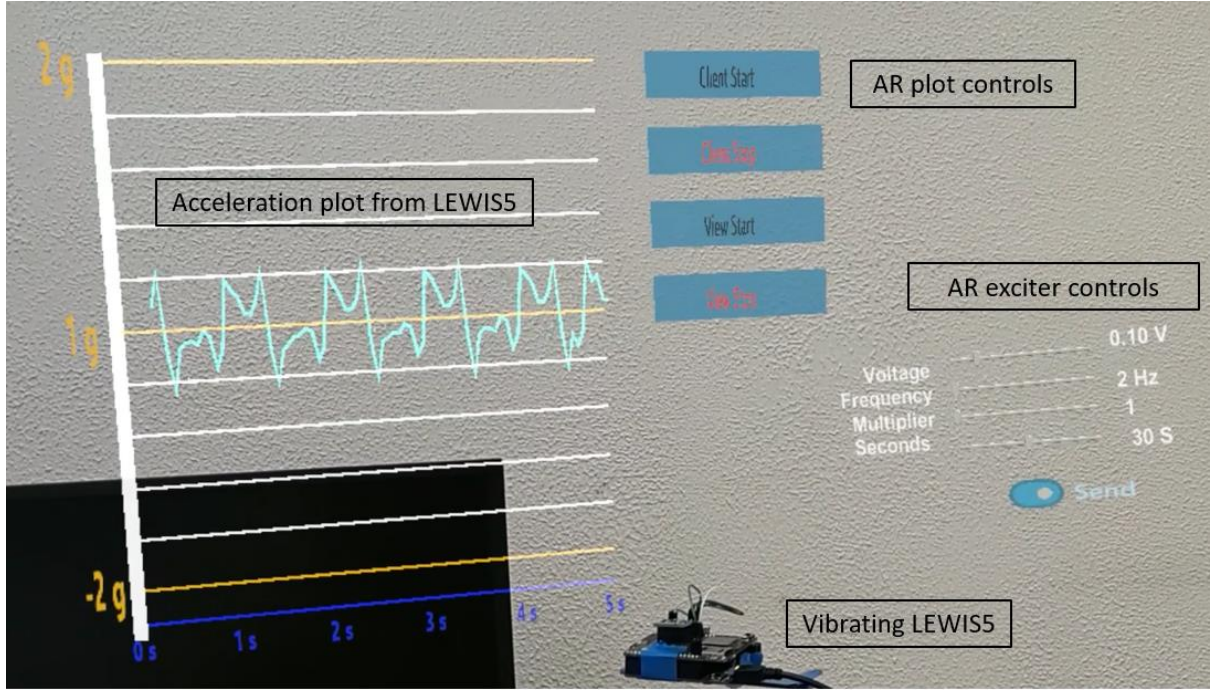

**Fig. 5** Actuator control application view in HL2.

**Exciter control time delay results**
The actuator control application was tested to determine the time delay in actuation. The time delay in the program was investigated using video analysis. Equation 3 calculates the frames between the initial sensor acceleration and the recorded response to approximate the time delay. The HoloLens camera records 1080p30 video. With the known value of the video framerate in FPS the time delay of the application can be calculated using the following formula.

$$(Frame_1 - Frame_0) * \frac{1}{Video\ FPS} = Time\ Delay \quad (3)$$

Frame$_0$ was designated as the time of initial acceleration by the sensor. Thus, Frame$_1$ designates the response plotted on the graph in the user's view. The difference in frames was recorded and the time delay was calculated with Equation 1. The results were processed in MATLAB and reported in scatter plot format in Figures 6 and 7. Figure 6 plots the time delay in seconds recorded for each of the 14 trials conducted with the standard method of using a function generator for input to the exciter. Figure 7 shows the time delay in seconds recorded for each trial with the AR control method. The time delay in exciter response with AR is an average of 0.37 seconds between the moment "send" is pressed and the exciter moves. The delay using the function generator comes out to an average of 0.20 seconds for the 14 trials.

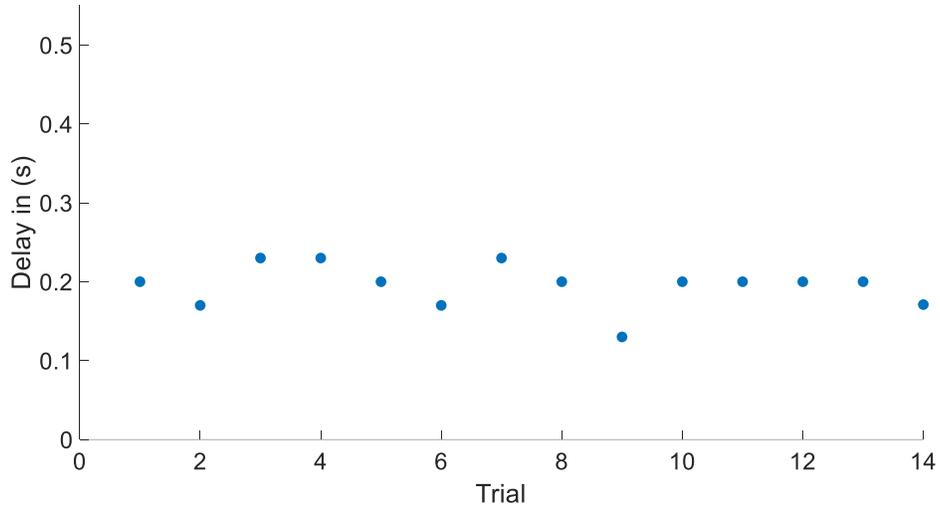

**Fig. 6** Experimental results of exciter control without AR.

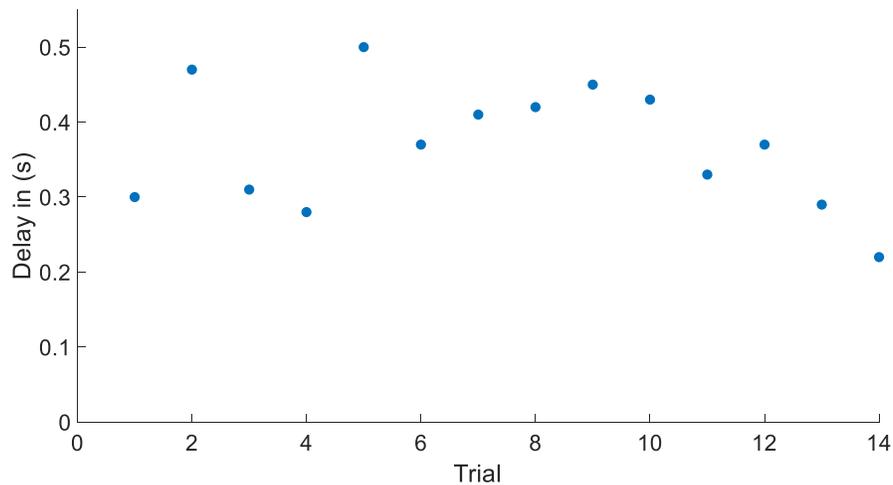

**Fig. 7** Experimental results of exciter control with AR.

This experimental result serves as a first-step investigation into AR control methods. While the results show that the traditional method of control is about 0.17 seconds, this result is better than expected considering the delay introduced in a network connection. By testing the time delay in control with and without AR a quantitative comparison can be drawn between the different control methods.

**Result of interface to control robot arm position**
The result of the position control application is an interface that allows novices and experts alike to adjust the position of a robotic arm with basic controls. The interface simply consists of a button to send the position vector between the two holograms, open and close gripper buttons, and a button to reset the robot to a rest position. As shown on the right the user moves the origin orb to the base of the robot and the target orb is set to a desired position. The application automatically connects to the database upon opening the application, so all that is necessary is selecting "send" to store the position.

**Result of interface to control robot movement**

The frequency control application is developed to inform the human of sensor data while also displaying the frequency of the human's own movement. The goal is for the human to move their hand at a rate desired for the robot arm, where the robot can then perform operations in a dynamic setting by counteracting undesirable movement. The full view of the interface is shown in Figure 8. The application interface consists of the plot of sensor data where the frequency of this response is updated in the bottom right corner of the graph. The same buttons from the actuator control application are included to connect to the server and begin viewing the data. The frequency updates every 2.67 seconds in the user's view and this frequency depends on the movement of the handheld LEWIS5 highlighted in blue. Boxed in red is the display of the active database seen on a laptop screen which updates with each new frequency value.

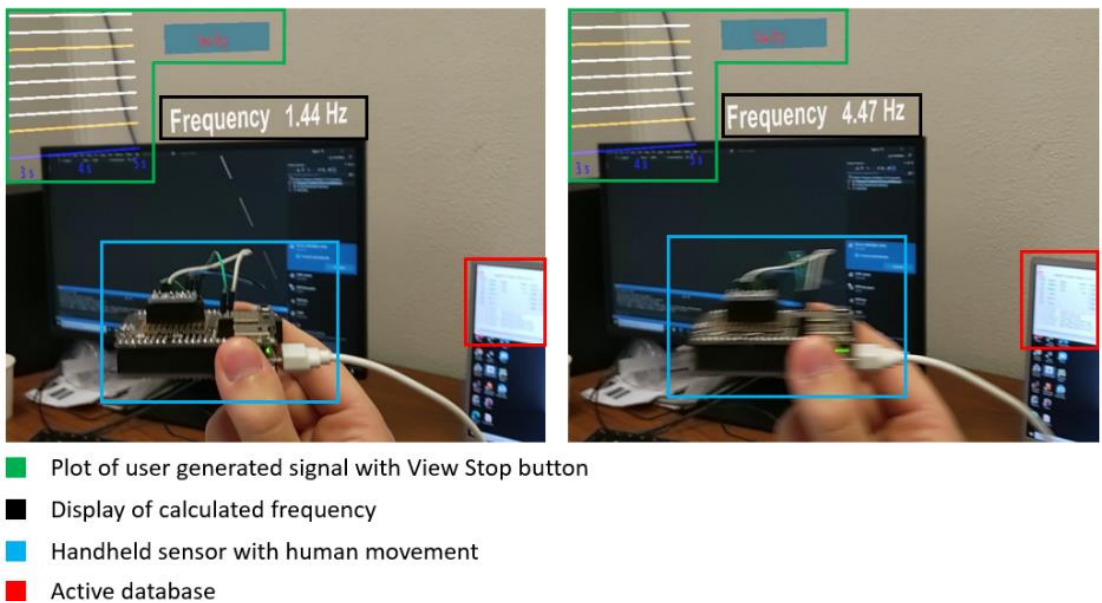

**Fig. 8** Components of robot frequency control as seen in HL2 view.

**Robot control experiment – controller versus AR results**

To test the viability of the position control application for the robot an experiment was designed to quantify the time it takes to deploy a sensor with the robot arm with and without AR. For the first session in the experiment, a novice user, and an expert user attempt to move the arm to a specific vertical position. The novice user is defined as a subject with zero experience with control of the Gen3 and AR, and the expert user is defined as a subject with over six months of experience. This is demonstrated in Figure 9a where the AR view of target and origin is shown, and Figure 9b which shows the robot reaching its target.

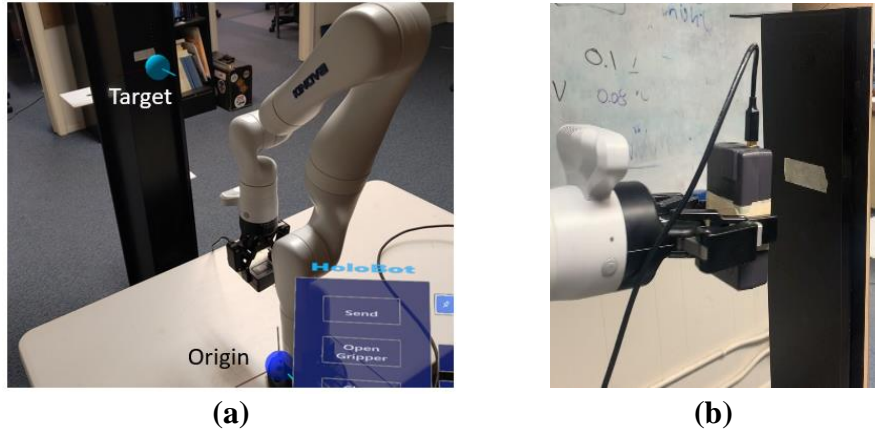

(a)           (b)

**Fig. 9** Implementation of position control. **(a)** Position control application developed deployed in HL2 (as seen in HL2 view); **(b)** demonstration by attaching magnetic senor box to target.

The first session was designed as a first-step investigation into novice and expert control with and without AR and serves as an introduction for the novice. For the second session in the experiment, a target position on a metal structure is marked with a strip of tape, and a sensor box with magnetic attachment is held by the robotic arm and placed on the structure. A successful attempt is defined as attaching the sensor to the line, otherwise the result is not used. In both sessions the results of control with a physical controller are compared to AR. The process starts with the robot at rest position for each test. The user locates the target and navigates to the target with the controller or defines the target with the hologram in AR. A sketch of the controller trajectory is shown in Figure 10 based on the z-axis of the robot-held sensor. With the controller the user begins moving the robot from rest forward in the direction of the target (A). The user manually adjusts the joint angles of the robot using the controller until the orientation of the sensor magnet is correct (B). Once adjusted to be in line with the target the final forward movement is conducted until the magnet attaches to the structure (C).

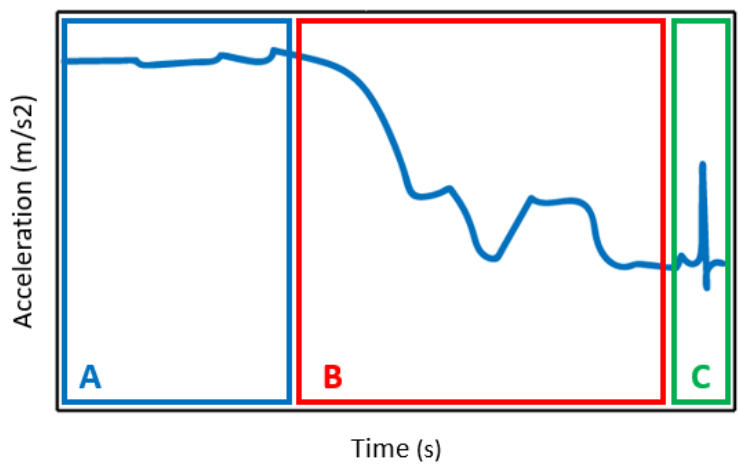

**Fig. 10** Trajectory of robot-held sensor axis when manipulated by traditional controller.

The time between the start of control and target position success was measured and recorded for five tests. The user starts movement by manual control and time is recorded based on the user's cognition for the first session and the time of attachment for the second session. With AR the application is opened in the user's view and the user defines the robot origin and target position with the corresponding holograms. For the first session time is recorded manually and for the second session robot-held sensor data is recorded to measure time and demonstrate robot movement. For AR the total time includes the time to define the target position in the application, and this is added to the time it takes for the MATLAB code to move the arm to the defined target. AR is not as perfect as the graph suggests but as soon as the first vibration is detected the data is cut. The reported results include a time comparison between novice and expert for each session, and the average time for each control mode is calculated and reported. Figures 11 displays the data collected from the robot-held sensor for the first manual control tests. The period at the beginning of the response is longer without AR as the human is always slower to adjust the robot orientation than the code run from AR. Figure 12 shows the novice's improvement from first to second test, and the final three tests are reported in Figure 13.

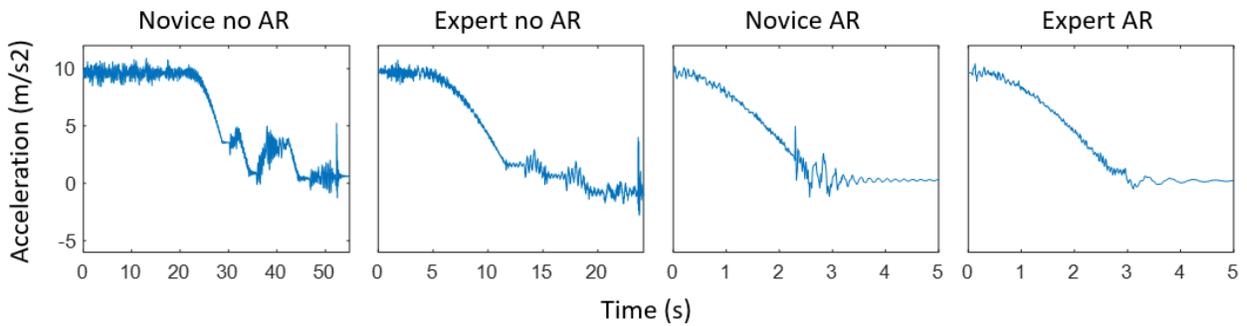

**Fig. 11** Time history of Test 1 moving sensor to target with robot.

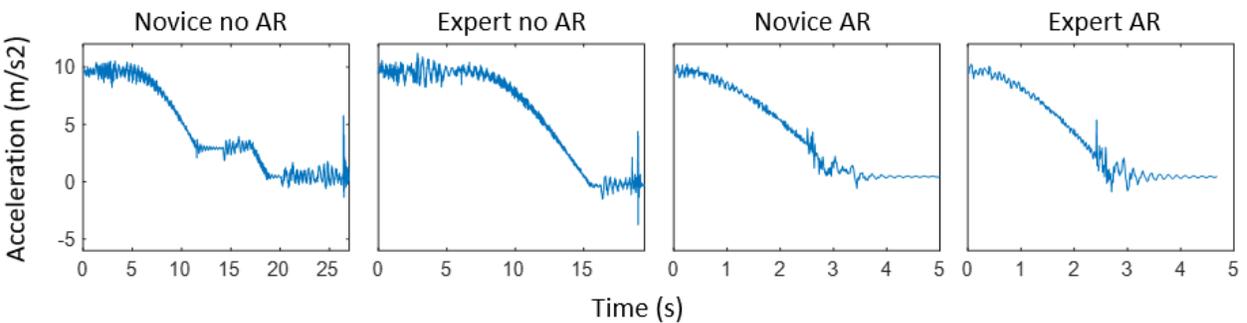

**Fig. 12** Time history of Test 2 – subject's improvement in moving sensor to target with robot.

The results show improvement for both operators between Test 1 and 2 and show a clear difference between novice and expert. The expert is nearly twice as fast in the first test, having needed less time to re-train due to experience. The novice improves by nearly 25 seconds but is still slower than the expert. The final three tests are plotted in Figure 13. Here the time without AR is normalized to a 20 second period for each plot. The time to move with AR is unchanged for all four time histories as the robot consistently executes the command at the same rate.

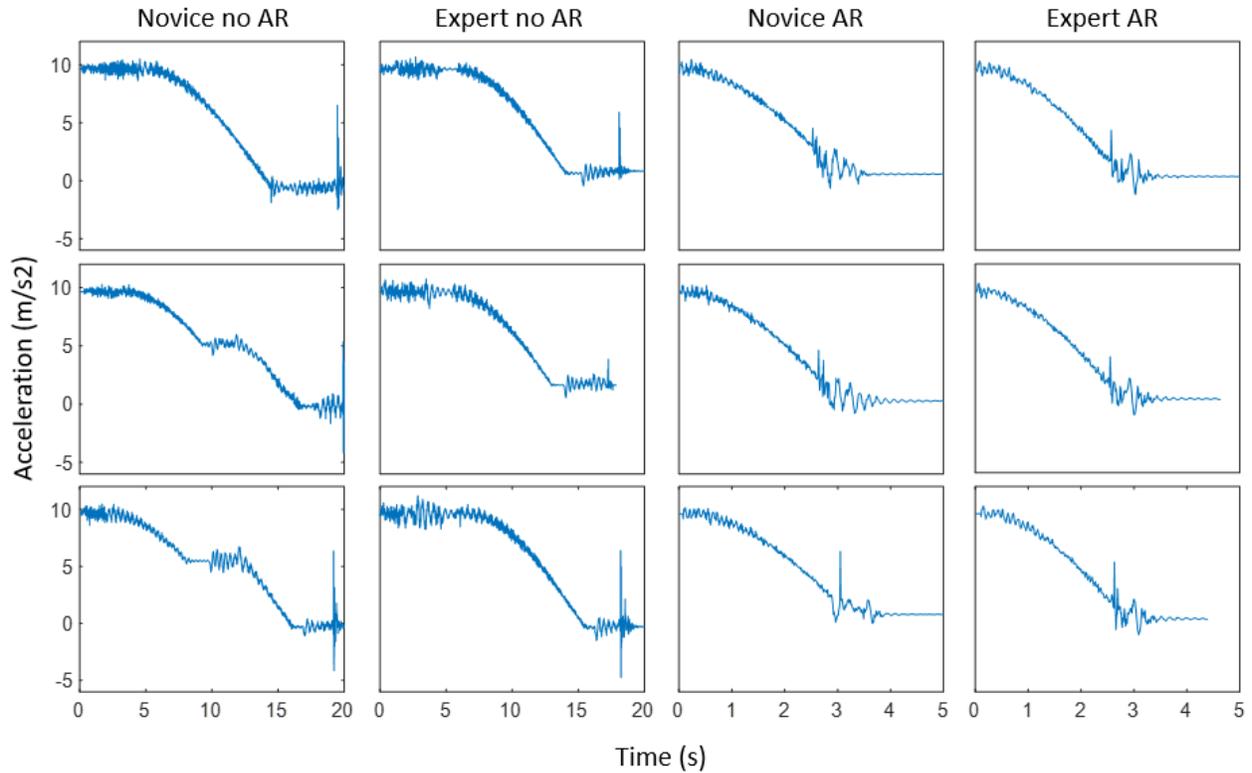

**Fig. 13** Time histories Test 3-5 moving sensor to target with robot.

For Figures 14 and 15, the results for controller are plotted from first excitation to 0.5 seconds after the peak that indicates the sensor contacted the target surface. Figure 14 demonstrates a low-quality result where the novice and expert need to re-train themselves on the correct control path. Contact with the target surface is noticeably quieter with AR in the expert's attempt versus the novice. Figure 15 shows the next step in the experiment where both subjects improve in the time it takes to reach the target.

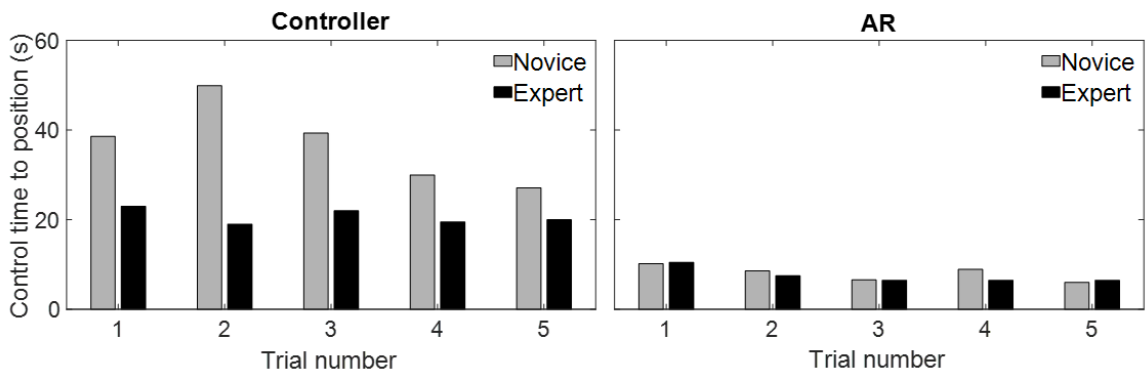

**Fig. 14** Total time to control for first session with and without AR.

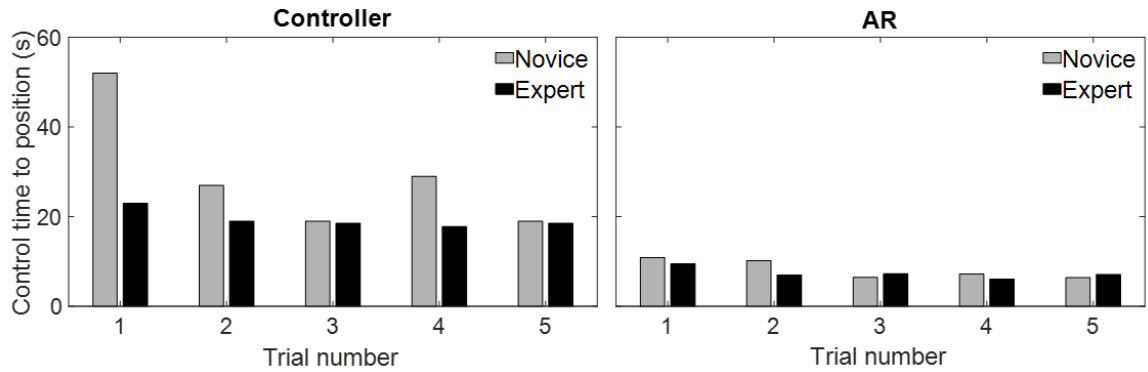

**Fig. 15** Total time to control for second session with and without AR.

The results show a significant difference between the untrained subject and the trained subject when controlling the arm with the physical controller. On average with the controller the expert was nearly 10 seconds faster than the novice. However, with AR the two subjects were very similar in the time to deploy the sensor. Session two repeated this experiment, except the target position on a metal structure was marked with a strip of tape and a sensor box with magnetic attachment was held by the robotic arm and placed on the structure. In this case, the novice improved much faster having already operated the robot in the first session. As was seen in the first session, the expert succeeded at a faster rate than the novice for the five tests, but the novice was much closer in time. Additionally, apart from the first trial, both novice and expert improved in time for the second session tests. Similar results were shown with AR as well, where again the novice performed efficiently from the start and was able to closely match the time of the expert.

**Robot frequency of movement experiment results**
The next experiment was designed to quantify the human's ability to follow platform movement represented by an actuator with and without the aid of the frequency control AR application. The goal of the human was to create a consistent, accurate frequency without breaching an initial limit of 2 Hz. The second part of the experiment quantifies how accurately the robot matches the frequency of the platform movement. Preliminary results testing human control with AR have reported that humans can best match frequency at under 3 Hz and improve when monitoring data in AR [23]. This analysis also reported decreasing error when attempting to match higher frequencies. Therefore, the frequencies selected for this experiment range should range from 0-2 Hz, the maximum speed for the robot. Four values are selected at 0.5 Hz, 1 Hz, 1.5 Hz, and 1.9 Hz, where it is hypothesized that the human can match the frequency by moving themselves. An actuator is used as a reference frequency to represent a moving base. The user follows the movement of the actuator as closely as possible to replicate the response. This value is calculated by the sensor and the sensor acceleration plot and frequency calculation is shown in the user's view.

The first part of the experiment runs for approximately one minute and the last 18 values are selected for the reported results, as for the first few values the user is adjusting to the setup. The test is repeated for each frequency for a total of four tests. The error between human and reference is induced by a combination of human error and error in the FFT calculation. Table 1 reports each of the 18 values generated by the user with and without AR.

**Table 1.** Results of human frequency following a reference frequency.

| Actuator | 0.5 Hz | | 1 Hz | | 1.5 Hz | | 1.9 Hz | |
|---|---|---|---|---|---|---|---|---|
| User | No AR | AR | No AR | AR | No AR | AR | No AR | AR |
| 1 | 0.53 | 0.55 | 1.05 | 0.97 | 1.64 | 1.52 | 1.91 | 1.92 |
| 2 | 0.42 | 0.58 | 1.04 | 1.03 | 1.51 | 1.55 | 1.73 | 1.96 |
| 3 | 0.69 | 0.41 | 0.77 | 1.05 | 1.55 | 1.54 | 1.89 | 1.99 |
| 4 | 0.32 | 0.64 | 0.81 | 1.00 | 1.55 | 1.60 | 1.93 | 1.88 |
| 5 | 0.43 | 0.55 | 1.02 | 1.02 | 1.71 | 1.49 | 2.00 | 1.94 |
| 6 | 0.61 | 0.24 | 1.04 | 1.07 | 1.81 | 1.50 | 2.09 | 1.91 |
| 7 | 0.5 | 0.55 | 0.91 | 0.88 | 1.53 | 1.55 | 1.95 | 1.99 |
| 8 | 0.47 | 0.38 | 0.97 | 0.97 | 1.50 | 1.49 | 1.90 | 1.94 |
| 9 | 0.3 | 0.38 | 1.12 | 0.99 | 1.50 | 1.57 | 1.98 | 1.94 |
| 10 | 0.21 | 0.45 | 1.00 | 1.06 | 1.50 | 1.52 | 1.89 | 1.96 |
| 11 | 0.65 | 0.53 | 1.02 | 1.05 | 1.51 | 1.55 | 1.97 | 1.85 |
| 12 | 0.44 | 0.52 | 1.10 | 1.03 | 1.53 | 1.55 | 2.08 | 1.98 |
| 13 | 0.3 | 0.51 | 1.03 | 1.00 | 1.59 | 1.49 | 1.71 | 1.91 |
| 14 | 0.47 | 0.54 | 1.05 | 1.02 | 1.51 | 1.53 | 1.9 | 1.95 |
| 15 | 0.49 | 0.53 | 0.94 | 1.07 | 1.44 | 1.42 | 1.93 | 1.89 |
| 16 | 0.34 | 0.54 | 0.94 | 0.88 | 1.58 | 1.51 | 1.96 | 1.95 |
| 17 | 0.46 | 0.48 | 0.92 | 0.97 | 1.54 | 1.53 | 2.01 | 1.85 |
| 18 | 0.42 | 0.52 | 1.10 | 0.99 | 1.56 | 1.54 | 2.06 | 1.93 |
| Average | 0.447 | 0.494 | 0.991 | 1.003 | 1.559 | 1.525 | 1.938 | 1.930 |
| Std. Dev. | 0.122 | 0.095 | 0.092 | 0.054 | 0.084 | 0.037 | 0.098 | 0.041 |

As shown in by the results in Table 1, the user creates a frequency that is much more consistent with the aid of AR than without. Monitoring the frequency also helps the human ensure that the threshold is not crossed. Without the knowledge imparted by visualization in AR, the user failed five times. To convey the trend in the consistency of the frequency generation the standard deviation in each set is reported in bar graph form in Figure 16.

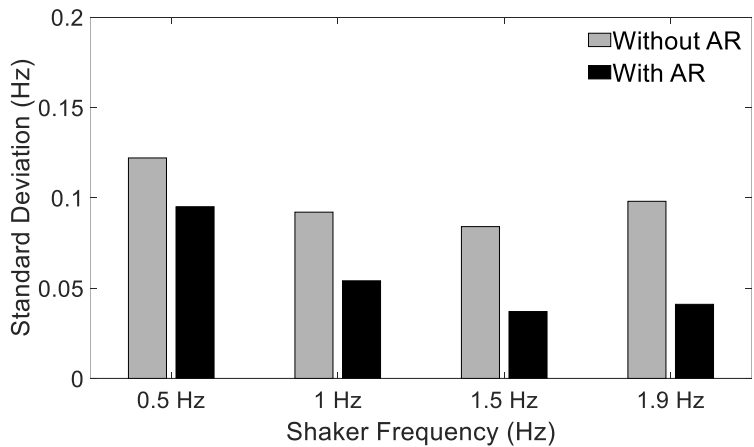

**Fig. 16** Results of standard deviation in frequencies generated by human.

The human had the most difficulty at the lowest frequency of 0.5 but performed similarly at the three higher frequencies. With and without AR the human followed the same trend of improvement between 1, 1.5, and 1.9 Hz however there is a significant improvement at each individual value with AR. Row 13 was selected as the values to run for the robot, where the arm holds the sensor box in its gripper to record the acceleration values. These values are 0.51, 1, 1.49, and 1.91 Hz. Auto-spectral density estimates were generated for each output and the time history comparisons are reported. The PSD curves were generated to determine the frequency of the robot movement to quantify accuracy. The actuator frequency was also checked since it cannot be assumed to be exact, and this value is plotted as a vertical line in each PSD shown below in Figure 17.

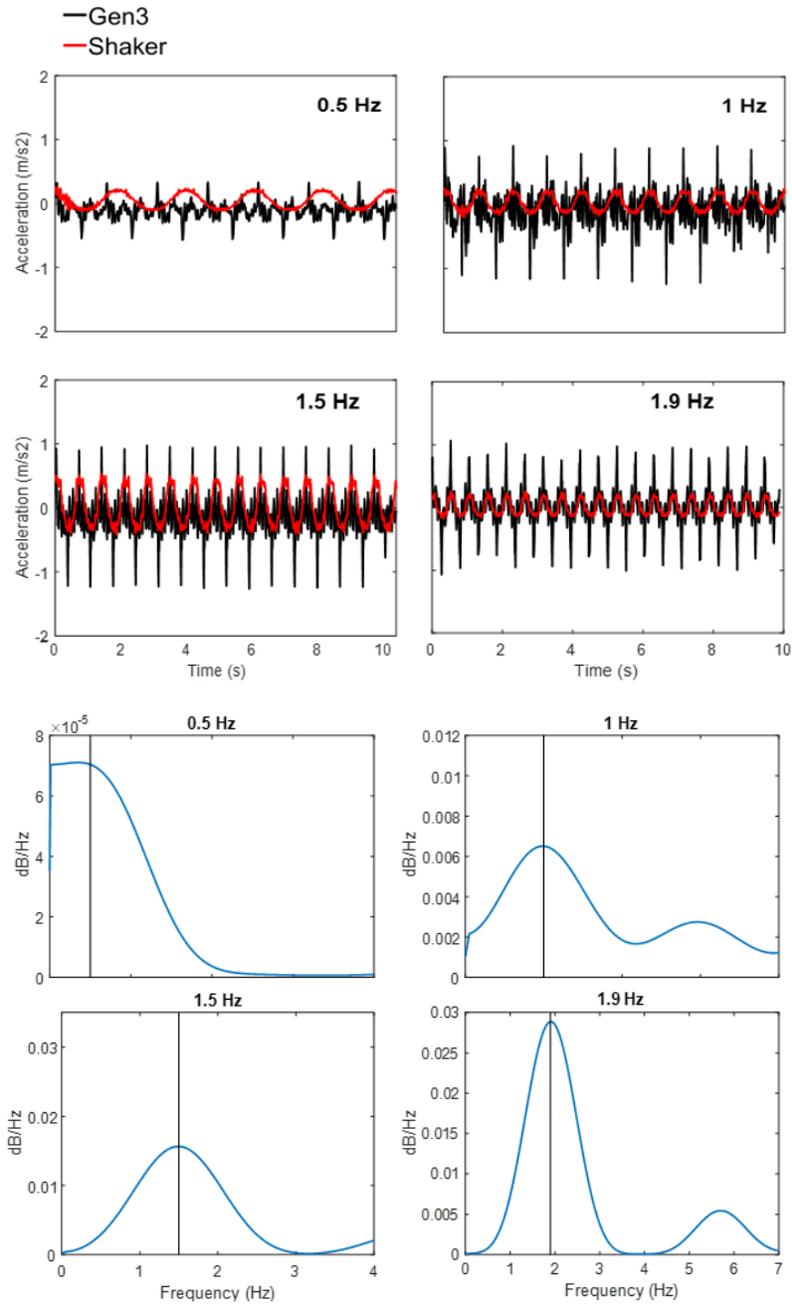

**Fig. 17** Results of robot movement compared to actuator response.

The Gen3 arm performs very well at matching the actuator frequency for the three highest values. However, at 0.5 Hz it does not run as efficiently. As demonstrated in the results, the robot movement does not generate a smooth curve that is in synch with the actuator. Rather, the resultant PSD gives 0.4 Hz as the result for 20% error. This is due to the joint movement of the robot, which performs shaky motion at such low frequency. At higher frequency the robot moves with more stable motion while matching the frequency with less than 2% error. As a result, it can be concluded that the robot would be best implemented for frequencies between 0.5-1.9 Hz, and environments with higher level of vibration should be avoided. Future work would see an implementation of the same framework where instead of running at an arbitrary time the robot is set to offset the movement exactly.

**Conclusions**
The research proposes using Augmented Reality applications to provide an interface for sensor feedback with control of actuators and robots for intelligent control of dynamics and experimentation. This method improves cognition by allowing the operator to (1) maintain awareness of the structure; (2) interact with data; (3) adjust conditions on-the-fly; and (4) operate actuators and robot mediation. An application was developed that plots sensor data in an AR interface. This approach was applied to an interface that includes voltage, frequency, and duration controls for vibration generation and an application for frequency control of a robotic arm. The new actuator control loop was tested and compared it to a traditional method of using a function generator by measuring time delay in control input. A robotic application for placement and stability was developed and tested for accuracy and validated as meaningful first-step approaches for robotic control solutions. The reported experiments prove that complex control is simplified with AR as the novice can compete with the expert in the total time to deploy a sensor. Frequency control is tested with AR where the user can better create a consistent frequency without breaching a defined limit, where the Gen3 can then match this frequency closely. These applications emphasize user awareness of the physical space by augmenting control and sensor feedback.

**Data availability**
The data evaluating the findings in this study are included in the article and its supplementary information. All data are available from the corresponding author upon reasonable request.

**Code availability**
All relevant code is available upon reasonable request from the corresponding author.

**Acknowledgements**
The financial support of this research is provided in part by the Air Force Research Laboratory (AFRL), grant number FA9453-18-2-0022; the New Mexico Space Grant Consortium (NMSGC), subaward number Q02151; the Office of Naval Research, grant number N00014-22-1-2638 and N00014-21-1-2169; Department of Energy (DOE), award number 0000525179; National Academy of Science Transportation Research Board (TRB) Rail SAFETY IDEA Program Project RS-43, project number 163420-0399; and the Federal Railway Administration (FRA) BAA project number FR20RPD34000000006. The authors would like to extend thanks to Marlan Ball and Jack Hanson for their assistance with the software development.

**Competing interests**
The authors declare no competing interests.

# References

1. Morimoto, R. (2013). A socio-economic analysis of Smart Infrastructure sensor technology. *Transportation Research Part C: Emerging Technologies, 31*, 18-29. Doi:10.1016/j.trc.2013.02.015
2. A. Sabato, C. Niezrecki and G. Fortino,""Wireless MEMS-Based Accelerometer Sensor Boards for Structural Vibration Monitoring: A Review"" in *IEEE Sensors Journal*, vol. 17, no. 2, pp. 226-235, 15 Jan.15, 2017, doi: 10.1109/JSEN.2016.2630008.
3. Li R-J, Lei Y-J, Chang Z-X, Zhang L-S, Fan K-C. Development of a High-Sensitivity Optical Accelerometer for Low-Frequency Vibration Measurement. *Sensors*. 2018; 18(9):2910. https://doi.org/10.3390/s18092910
4. Wang, S., Wei, X., Zhao, Y., Jiang, Z., & Shen, Y. (2018). A MEMS resonant accelerometer for low-frequency vibration detection. *Sensors and Actuators A: Physical*, *283*, 151–158. https://doi.org/10.1016/j.sna.2018.09.055
5. Abruzzese, D., Angelaccio, M., Giuliano, R., Miccoli, L., & Vari, A. (2009). Monitoring and vibration risk assessment in cultural heritage via Wireless Sensors Network. *2009 2nd Conference on Human System Interactions*. Doi:10.1109/hsi.2009.5091040
6. Saadatzi, M., Saadatzi, M. N., Tavaf, V., & Banerjee, S. (2018)."”AEVE 3D: Acousto Electrodynamic Three-Dimensional Vibration Exciter for Engineering Testing”” in *IEEE/ASME Transactions on Mechatronics*, vol. 23, no. 4, pp. 1897-1906, doi: 10.1109/TMECH.2018.2841011.
7. Čala, M. (2015)""Control system for a small electrodynamic exciter"" Proceedings of the 2015 16th International Carpathian Control Conference (ICCC), pp. 64-68, doi: 10.1109/CarpathianCC.2015.7145047.
8. Popa, D. O., Stephanou, H. E., Helm, C., & Sanderson, A. C. (2004). Robotic deployment of sensor networks using potential fields. *IEEE International Conference on Robotics and Automation, 2004. Proceedings. ICRA '04. 2004*. https://doi.org/10.1109/robot.2004.1307221
9. Prusaczyk, P., Kaczmarek, W., Panasiuk, J., & Besseghieur, K. (2019). Integration of robotic arm and vision system with processing software using TCP/IP protocol in industrial sorting application. *AIP Conference Proceedings*. https://doi.org/10.1063/1.5092035.
10. Palacios, R.H. (2015). Robotic Arm Manipulation Laboratory WithF a Six Degree of Freedom JACO Arm. [Thesis]. Naval Postgraduate School.
11. Wisanuvej, P., Liu, J., Chen, C. & Yang, G.H. (2014). Blind collision detection and obstacle haracterization using a compliant robotic arm. Proceedings– IEEE International Conference on Robotics and Automation. 2249-2254. 10.1109/ICRA.2014.6907170.
12. Manring, L., Pederson, J., Potts, D., Boardman, B., Mascareñas, D., Harden, T., & Cattaneo, A. (2020). Augmented reality for interactive robot control. In *Special Topics in Structural Dynamics & Experimental Techniques, Volume 5* (pp. 11-18). Springer, Cham.
13. Horbst, J. (2020). GoHolo. [Thesis]. University of Applied Sciences Technikum Wien.
14. Chacko, S. M., Granado, A., RajKumar, A., & Kapila, V. (2020, October). An Augmented Reality Spatial Referencing System for Mobile Robots. In *2020 IEEE/RSJ International Conference on Intelligent Robots and Systems (IROS)* (pp. 4446-4452). IEEE.
15. Moreu, F., Lippitt, C., Maharjan, D., Aguero, M., & Yuan, X. (2019). Augmented Reality Enhancing the Inspections of Transportation Infrastructure: Research, Education, and Industry Implementation. Retrieved from https://digitalcommons.lsu.edu/transet_data/57.
16. Ashour, Z., & Yan, W. (2020). BIM-Powered Augmented Reality for Advancing Human-Building Interaction.
17. Aguero, M., Maharjan, D., Rodriguez, M. D., Mascareñas, D. D., & Moreu, F. (2020). Design and Implementation of a Connection between Augmented Reality and Sensors. *Robotics, 9*(1), 3. Doi:10.3390/robotics9010003.
18. Younis, O., Al-Nuaimy, W., & Rowe, F. (2019). A hazard detection and tracking system for people with peripheral vision loss using smart glasses and augmented reality. *International Journal of Advanced Computer Science and Applications*, *10*(2).
19. Andersson, N., Argyrou, A., Nägele, F., Ubis, F., Campos, U. E., Zarate, M. O., & Wilterdink, R. (2016). AR-Enhanced Human-Robot-Interaction– Methodologies, Algorithms, Tools. *Procedia CIRP, 44*, 193-198. Doi:10.1016/j.procir.2016.03.022
20. MTS Systems. (n.d.). *SmartShaker™ with Integrated Power Amplifier*. The Modal Shop, Inc. https://www.modalshop.com/excitation/SmartShaker-with-Integrated-Power-Amplifier?ID=272.
21. Kuzhagaliyev, T. (2018, May 27). *TCP client in a UWP Unity app on HoloLens*. Foxy Panda. https://foxypanda.me/tcp-client-in-a-uwp-unity-app-on-hololens/.
22. Flick, J. (2020, September 23). Building a Graph. https://catlikecoding.com/unity/tutorials/basics/building-a-graph/.

23. Wyckoff, E., Ball, M., & Moreu, F. (2022). Reducing gaze distraction for real-time vibration monitoring using augmented reality. *Structural Control and Health Monitoring*. https://doi.org/10.1002/stc.3013.